# Extracting urban water by combining deep learning and Google Earth Engine

Yudie Wang, Zhiwei Li, Chao Zeng, Guisong Xia and Huanfeng Shen

*Abstract*—**Urban water is important for the urban ecosystem. Accurate and efficient detection of urban water with remote sensing data is of great significance for urban management and planning. In this paper, we proposed a new method to combine Google Earth Engine (GEE) with multiscale convolutional neural network (MSCNN) to extract urban water from Landsat images, which is summarized as offline training and online prediction (OTOP). That is, the training of MSCNN was completed offline, and the process of urban water extraction was implemented on GEE with the trained parameters of MSCNN. The OTOP can give full play to the respective advantages of GEE and CNN, and make the use of deep learning method on GEE more flexible. It can process available satellite images with high performance without data download and storage, and the overall performance of urban water extraction is also higher than that of the modified normalized difference water index (MNDWI) and random forest. The mean kappa, F1-score and intersection over union (IoU) of urban water extraction with the OTOP in Changchun, Wuhan, Kunming and Guangzhou reached 0.924, 0.930 and 0.869, respectively. The results of the extended validation in the other major cities of China also show that the OTOP is robust and can be used to extract different types of urban water, which benefits from the structural design and training of the MSCNN. Therefore, the OTOP is especially suitable for the study of large-scale and long-term urban water change detection in the background of urbanization.**

## I. INTRODUCTION

URBAN water is a significant part of the urban ecosystem and plays an important role in human life and urban economic development, such as water supply, flood control, tourism and urban heat island regulation [1], [2]. Urban waters include rivers flowing through urban areas, natural or man-made lakes, ponds, reservoirs and so on. However, in recent decades, due to the impact of human activities and global climate change, land use/cover in urban areas has undergone severe changes, resulting in dramatic changes in the distribution and abundance of urban water [3], [4]. This not only hinders the sustainable development of urban water resources, but also affects the normal function of the urban ecosystems [5]. Therefore, an objective and accurate understanding of the spatial-temporal distribution characteristics of urban water is essential for urban planning and development.

Satellite remote sensing technology has been widely used in the mapping of water, because of its wide range of observations, relatively low cost, etc. [6], [7]. Previous studies have proposed many methods for water extraction from satellite images. The water index methods are extensively used because of its easy implementation and high computational efficiency, such as the normalized difference water index (NDWI) [8], the modified normalized difference water index (MNDWI) [9] and automatic water extraction index (AWEI) [10]. These water indices can enhance water information based on the spectral reflectance characteristics of water in visible and infrared bands. However, it is difficult to determine the optimal thresholds of different images for effectively distinguishing between water and non-water, because the spectral characteristics of water vary spatially and temporally [11]. And for urban water extraction, these indices also have the problem of easily confusing urban water with low-albedo non-water surfaces, such as asphalt roads, building shadows, etc. [10].

Many machine learning methods are also used to extract urban water from satellite images to improve the classification accuracy, such as maximum likelihood (ML) [12], [13], support vector machine (SVM) [14]-[16] and random forest [17]-[19], etc. Most of these methods can effectively learn the characteristics of water and distinguish it from other objects in the case of accurate training samples. But in addition to relying on high-quality training data, these methods also need to select appropriate combinations of features, which may include spectral, textural and shape features, especially in urban areas. Feature selection will directly affect the stability of prediction, and the design of features is not only time-consuming and laborious, but also needs rich prior experience to make up for the shortcomings of data mining.

As a subset of machine learning, deep learning, especially the convolutional neural network (CNN), has also demonstrated its effectiveness in image processing in recent research [20]-[22]. The greatest advantage of CNNs is the strong ability to extract

Corresponding author: Huanfeng Shen.

Y. Wang is with the State Key Laboratory for Information Engineering in Surveying, Mapping, and Remote Sensing (LIESMARS), Wuhan University, Wuhan 430079, China (e-mail: ydiewang@whu.edu.cn).

Z. Li is with School of Resource and Environmental Sciences, Wuhan University, Wuhan 430079, China (e-mail: lizw@whu.edu.cn).

C. Zeng is with School of Resource and Environmental Sciences, Wuhan University, Wuhan 430079, China (e-mail: zengchaozc@hotmail.com).

G. Xia is with the State Key Laboratory of Information Engineering in Surveying, Mapping and Remote Sensing (LIESMARS), Wuhan University, Wuhan 430079, China (e-mail: guisong.xia@whu.edu.cn).

H. Shen is with the School of Resource and Environmental Sciences, Wuhan University, Wuhan 430079, China, and also with the Collaborative Innovation Center of Geospatial Technology, Wuhan University, Wuhan 430079, China (e-mail: shenhf@whu.edu.cn).

multiscale and multilevel features. Some scholars have used CNNs to extract water from different types of images, and the results showed that CNNs could accurately distinguish between water and ice/snow, cloud shadows and terrain shadows without additional auxiliary materials [23]-[26]. However, most of these studies were small-scale or short-term. For large-scale and long-time urban water extraction, the download and storage of massive satellite images and the requirement for high computational performance are unavoidable challenges.

Google Earth Engine (GEE), as a cloud platform dedicated to geographic data processing and analysis, can solve those problems well. GEE provides massive global geospatial data and many excellent image processing algorithms, and all processing is parallel [27], [28]. These enable researchers to perform large-scale and long-term analysis with minimal cost and equipment [29]-[33], including surface water mapping [34]-[37]. However, up to now, the function of deep learning on GEE is still developing. The existing research on the combination of GEE and deep learning was mainly to use GEE as the data source or a platform for creating training and test datasets, whereas the training and testing of the deep learning models were based on other platforms [38]-[40]. It does not fully demonstrate the advantages of GEE designed specifically for managing big data.

Therefore, we proposed a new method to combine GEE with CNN model to detect urban water. A completed CNN model called multiscale convolutional neural network (MSCNN) was trained offline with the selected Landsat images and the corresponding water masks. Then the trained model parameters were uploaded to GEE and used to simulate the process of MSCNN's urban water detection to complete the urban water extraction of Landsat images on GEE. The proposed framework is summarized as offline training and online prediction (OTOP). The goal of the OTOP method is to provide a more flexible way to combine the preferred deep learning model and GEE. It can make full use of the high-precision advantage of CNN and the massive data and powerful computing performance of GEE, so as to achieve efficient and accurate urban water detection.

## II. Experimental Data

Based on the advantages of high spatial resolution (30m) and long observation records (1972-present), Landsat data has become one of the most commonly used data types for monitoring long-term changes of water [41]-[46], which was also used in this study. The Landsat surface reflectance (SR) data on GEE were selected to ensure the consistency of data. The Landsat TM (Thematic Mapper) and ETM+ (Enhanced Thematic Mapper Plus) SR data were generated from Landsat standard Level 1 Terrain-corrected (L1T) images in USGS with the Landsat Ecosystem Disturbance Adaptive Processing System (LEDAPS) algorithm [47], and the Landsat OLI (Operational Land Imager) SR data was generated in USGS with the Landsat Surface Reflectance Code (LaSRC) algorithm [48].

The characteristics of urban water are different due to the diversity of urban terrains and urban development plans. In order to ensure the strong generality and robustness of the OTOP method, the selection of training data needed to consider as many cases as possible. In this paper, 36 Landsat images covering 36 major cities in China, most of which are provincial capitals and municipalities, were chosen as experimental data. These 36 cities basically represented the situations of urban water under different climates, topographies and development levels in China. The 36 images we used were acquired at different times and came from three different sensors, Landsat TM, ETM+, and OLI. These images also covered many different surface features. In addition, cloudy images were also included in the experimental data, because the effects of cloud and cloud shadows were inevitable in the real world. In order to increase the differentiation between water and other objects in urban areas, the sample size of urban areas was appropriately increased in the training data. Therefore, the CNN model trained by the dataset selected under the multiple considerations can be used for the extraction of urban water in different cities of China at different times. The spatial distributions and acquisition times of the selected images are shown in Fig. 1 and Table I.

The water masks of the selected 36 Landsat images were manually drawn by referring to the original images and the high-resolution images of the same periods on Google Earth. The mapping of water masks followed uniform standards, for example: (1) the water with less than 4 pixels was not drawn. (2) The slender water with a width of less than 2 pixels was not easy to determine the boundary, so it was ignored. (3) According to actual experience, the mixed pixels located at the edge of water areas were identified as water if the water features were obvious, otherwise as non-water, etc. Finally, the manually labeled reference water masks were created by setting the pixel values of water and non-water to 1 and 0, respectively.

For the accuracy verification of the OTOP method for urban water extraction, we mainly chose the central urban areas of Changchun, Wuhan, Kunming and Guangzhou as test areas. The extents of the central urban areas were determined by the latest urban planning of each city. And Changchun is located in the Northeast Plain of China and the water in the territory is mainly tributaries of rivers. Wuhan, located in the Jianghan Plain in central China, is dotted with nearly hundred lakes in the territory which is known as the "city of hundred lakes". Kunming is located in the middle of the Yunnan-Guizhou Plateau in southwestern China, and most of the waters are plateau lakes and reservoirs. Guangzhou is a coastal city of China, located in the hilly area, with developed river system and vast water area. There are basically all types of urban water with different depths, turbidities and surface morphologies in the four cities, and the spectral features of the built-up areas of the four cities are also different, covering most of the major challenges affecting the extraction of urban water, such as building shadows, low albedo roofs and roads. The images of the four cities used for testing were also from different sensors. Thus, the accuracy assessment of urban water extraction in these four cities can better reflect the ability of the OTOP method. In the experiment, the data of the other 32 cities were used for model training.

TABLE I
THE WRS-2 PATH/ROW, THE ACQUISITION TIME AND CORRESPONDING CITIES OF LANDSAT TM, ETM+, OLI IMAGES USED IN THIS STUDY.

| Sensor | Path/Row | Date | City | Sensor | Path/Row | Date | City | Sensor | Path/Row | Date | City |
|---|---|---|---|---|---|---|---|---|---|---|---|
| ETM+ | 117/043 | 2001-03-06 | Taipei | TM | 120/038 | 2004-12-08 | Nanjing | ETM+ | 125/034 | 1999-11-20 | Taiyuan |
| TM | 118/028 | 2009-05-14 | Harbin | TM | 121/038 | 1995-09-02 | Hefei | TM | 125/044 | 1994-03-03 | Nanning |
| ETM+ | 118/029 | 2000-04-11 | Jilin | OLI | 121/040 | 2015-09-09 | Nanchang | OLI | 126/032 | 2017-02-05 | Hohhot |
| OLI | 118/030 | 2015-04-13 | Changchun | OLI | 122/033 | 2016-05-13 | Tianjin | TM | 127/036 | 1995-06-08 | Xian |
| TM | 118/038 | 2006-04-20 | Shanghai | OLI | 122/035 | 2013-11-29 | Jinan | TM | 127/042 | 2011-05-19 | Guiyang |
| TM | 118/039 | 1998-02-09 | Ningbo | TM | 122/044 | 2005-11-23 | Guangzhou | OLI | 128/039 | 2014-08-06 | Chongqing |
| TM | 119/031 | 1991-08-24 | Shenyang | ETM+ | 123/032 | 2001-06-04 | Beijing | TM | 129/033 | 2010-07-01 | Yinchuan |
| TM | 119/039 | 1999-10-01 | Hangzhou | TM | 123/039 | 1995-04-09 | Wuhan | ETM+ | 129/043 | 2000-02-20 | Kunming |
| TM | 119/042 | 2007-01-08 | Fuzhou | TM | 123/040 | 2004-09-24 | Changsha | TM | 131/035 | 2005-05-30 | Lanzhou |
| TM | 119/043 | 1997-01-28 | Xiamen | TM | 124/034 | 2010-10-02 | Shijiazhuang | TM | 132/035 | 1997-11-07 | Xining |
| TM | 120/033 | 2008-02-03 | Dalian | TM | 124/036 | 1992-10-16 | Zhengzhou | TM | 138/039 | 2003-11-18 | Lhasa |
| TM | 120/035 | 1996-05-08 | Qingdao | OLI | 124/046 | 2015-06-26 | Haikou | ETM+ | 142/030 | 2002-10-18 | Urumqi |

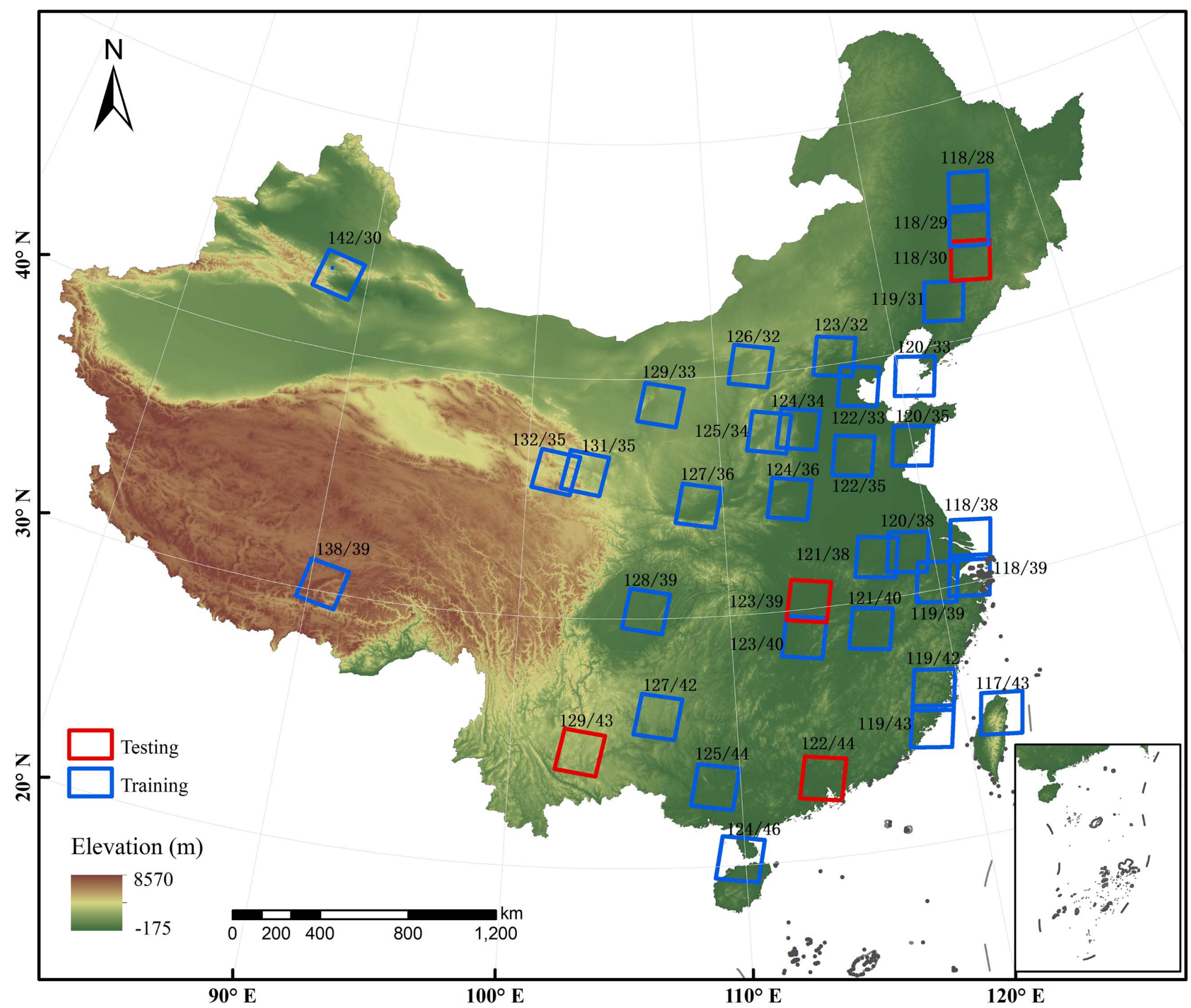

Fig. 1. The spatial distributions of experimental data in China. The blue and red boxes represent the spatial locations of the images used for training and testing in the accuracy assessment, respectively.

## III. METHODOLOGY

The flow chart of the proposed OTOP framework in this study is shown in Fig. 2. It is mainly divided into two parts: offline training with MSCNN and online prediction on GEE. And the details are described in the following sections.

### *A. Offline training*

#### *1)* Data preparation

The training data and test data consisted of Landsat TM/ETM+/OLI SR images and the corresponding water masks which were manually created. To better eliminate the interference of building shadows in water extraction, the training data included not only the images and water masks of 32 cities, but also the data of built-up areas specially clipped in these images.

The input of the network model was three visible bands, near infrared band and two short infrared bands of Landsat image. The pixel values of SR images were unified to the range of [0, 1] by dividing by 10,000 before the data was used for model training firstly. In the online prediction stage, the Landsat images should be processed with the same way as the training stage. Then, before training the model, a moving window was used to clip the large multispectral images and masks to pairs of non-overlapping training samples with a height and width of 256 × 256 pixels. Finally, there were more than 30,000 pairs of image blocks for model training.

*2)* Multiscale CNN model

As shown in Fig. 2, the CNN model used in the offline training, which is called MSCNN, is a simplified implementation of the Multiscale Convolutional Feature Fusion (MSCFF) model proposed by [49]. The MSCFF model was designed for cloud/cloud shadow detection of multisource remote sensing images, and the advantages of MSCFF were obvious compared with the traditional rule-based method and the existing deep learning model. So the MSCFF model was considered for urban water extraction. The complex encoder-decoder module of the MSCFF was simplified in the MSCNN model, under the comprehensive consideration of model accuracy and computational efficiency. In addition, the multiscale convolutional feature fusion module of the MSCFF was also used in the MSCNN model to further improve the accuracy of urban water detection by making full use of convolutional features of different scales.

The MSCNN model was mainly comprised of convolution operations at six different scales. At each scale, the image was processed by three multi-feature convolution operation layers and a single feature convolution operation layer which was used to extract higher-level feature information and reduce the channel of the feature maps. And besides the final scale, the last multi-feature convolution layer of each scale was followed by a pooling layer that reduced the spatial dimensions of the input data while ensuring translational invariance of feature extraction. The maximum pooling operation with 2 × 2 filters and a two-pixel stride was used, which returned the maximum value for each window. The six single feature maps of different scales were sampled to the same size as the input data, and the up-sampling results were aggregated as the input of the last convolutional layer. Such multiscale feature fusion could provide a broad description of the context of the given spatial locations [23]. The up-sampling operation in the MSCNN was directly implemented by bilinear interpolation. The output of the last convolutional layer of the network was processed by the softmax layer to output the class probability map of the water and non-water. The class with the highest probability would be regarded as the final classification result.

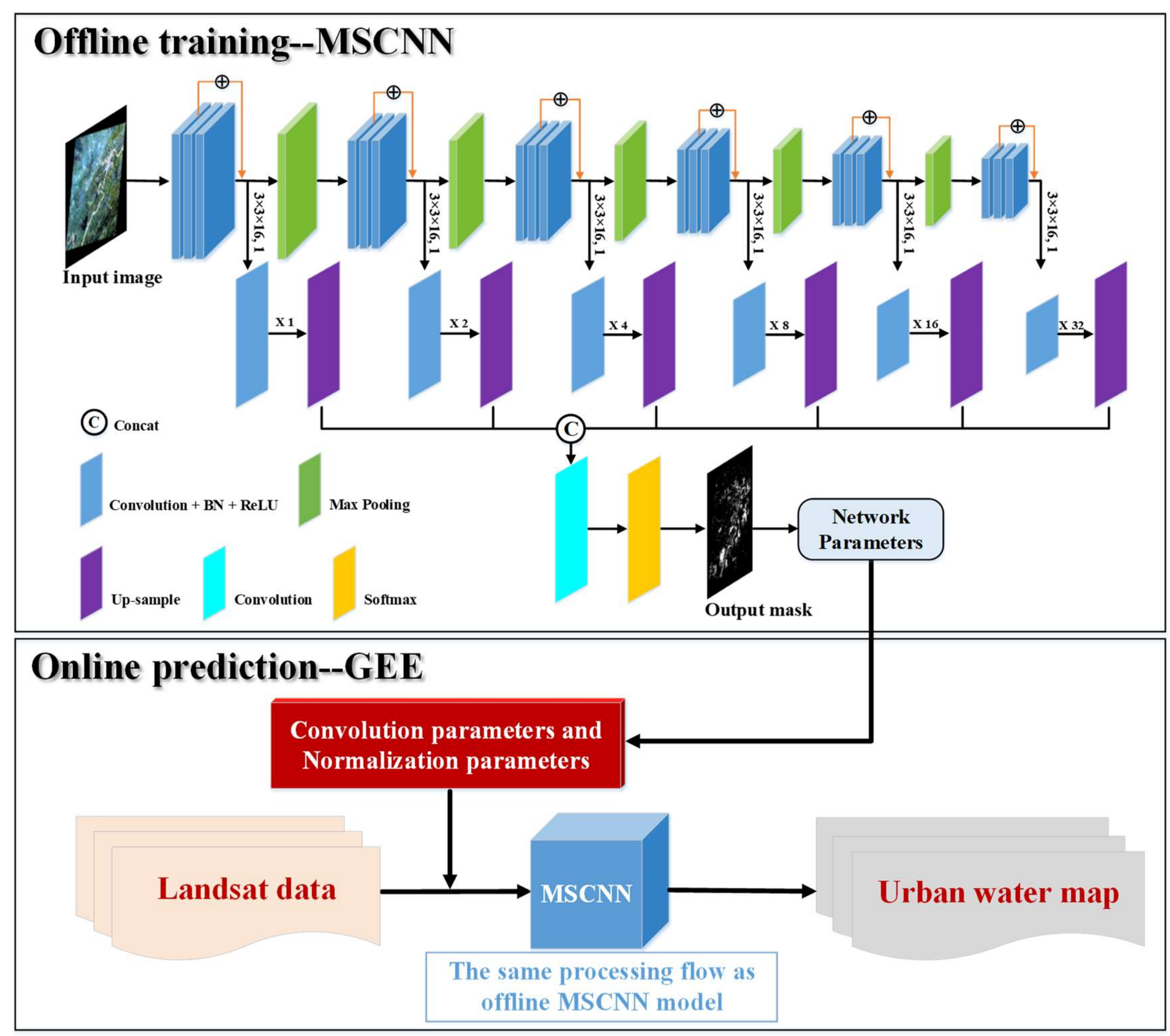

Fig. 2. The flow chart of the OTOP framework.

Moreover, in the MSCNN model, each convolutional layer, except for the last one, was followed by a batch normalization layer (BN) and a Rectified Linear Units (ReLU, f(x) = max (0, x)). The BN layer can speed up the training of the network and avoid the problem of gradient disappearance. ReLU, as a non-linear activation function, can introduce non-linear factors into the network and ensure that parameters can continue to converge. The convolutional kernel size of each convolutional layer was 3 × 3, and the output feature maps of each layer were kept the same size as the input data by padding zeros at the image borders in the convolution operation. The cross-entropy loss was used as the loss function in this model to measure the differences between the predicted and the expected results, which can be calculated as follows:

$$loss = -\frac{1}{N}\sum_{x=i}^{x=N}\left[ y_i \ln F\left(x_i, w\right) + \left(1-y_i\right) \ln\left(1 - F\left(x_i, w\right)\right)\right] \quad (1)$$

where $x_i$ is the input, $y_i$ is the expected output, $F(x_i, w)$ is the actual output, and $N$ is the number of training samples.

In the training process, the stochastic gradient descent (SGD) and back-propagation (BP) were used to update weight parameters $w$ until the optimal parameters were obtained by minimizing the loss function. The initial learning rate of the model was set to 0.1, and the learning rate followed the polynomial decay from 0.1 to 0 before the maximum number of iterations (200,000) was reached. Finally, the optimal weight parameters of the model were stored and then uploaded to GEE for online urban water detection.

### B. *Online prediction*

The most important step to extract the extents of urban water on GEE was to simulate the process of water detection of the offline MSCNN model with the uploaded trained parameters. As mentioned above, the training processes in the MSCNN model were mainly composed of convolutional operations, maximum pooling, up-sampling operations and basic arithmetic operations of images. So these basic operations were implemented one by one on GEE, and then integrated according to the framework of the MSCNN model to realize the extraction of urban water using the CNN model in the cloud platform.

The GEE platform provides many encapsulated basic algorithms while providing users with an online JavaScript API that allows them to create and run custom algorithms. It supports a variety of complex geospatial analysis, including image classification, change detection, time series analysis, etc. [50]. In addition, users can upload chart, raster or vector data to GEE for processing, and our research makes full use of the characteristic. The convolutional operations and image algebra operations of the MSCNN model can be implemented directly on GEE. The parameters of the convolutional kernels also support user-defined, so the kernel of each convolutional operation can be defined with the uploaded parameters of MSCNN. The process of maximum pooling is equivalent to down-sampling, and GEE provides reducer and reprojection operations to change image resolution. Similarly, the up-sampling process can also be directly achieved by reprojection and resampling (bilinear interpolation) on GEE. The results of the urban water detection code implemented by simulating MSCNN on GEE are the same as the results offline, which is the maximum probability class map. Based on the results, the final water extents can be determined.

## IV. Results

There were two parts to verify the feasibility and robustness of the OTOP method. The first part was the accuracy assessment of the method, and the predicted results of the OTOP were also compared with the urban water extraction results of other methods. The second part was to apply the OTOP method to other major cities in China and evaluate their accuracies. The specific results are shown in the following sections.

### A. *Accuracy assessment*

The OTOP method was compared with two typical methods, that is the traditional MNDWI method [9] and the random forest method [51], to prove the effectiveness of the combination of GEE and MSCNN in urban water extraction. The thresholds of MNDWI were set to the most suitable values which were determined artificially through multiple experiments. The algorithm of random forest method (RF) has been encapsulated on GEE platform already, which can be called directly. The six bands of Landsat images were selected as feature variables of the RF that were the same as the input of the OTOP. The training samples were also the same as the MSCNN model. The number of classification trees was determined to be 40 by comparative experiments, because the performance of the RF was not substantially improved when greater than 40. It's worth noting that both the MNDWI and the RF method were implemented on GEE.

#### *1)* Urban water extraction maps

Urban water extraction maps of the central urban areas of Changchun, Wuhan, Kunming and Guangzhou generated by MNDWI, RF and the OTOP method are shown in Fig. 3-6. On the whole, all methods could accurately extract evident and clear urban waters, such as large rivers, lakes, reservoirs, ponds. But compared with MNDWI and RF, the OTOP method performed better in the presence of the complex urban surface. Especially in the urban centers with dense buildings, the OTOP method could more accurately distinguish water from other non-water objects in the four test cities.

Visually, the traditional MNDWI was the most serious case of dividing the building shadows into waters in urban water extraction. There were different degrees of false detection in four cities, among which it was the most obvious in Guangzhou where high-rise buildings were dense (Fig. 7). It is difficult to distinguish the building shadows and low albedo objects from waters with a threshold because they had similar index values. The results of the RF had some improvement over those of MNDWI. But the RF also had the similar problem of false detection in urban areas. This indicated that the accuracy of the RF also depended on the selection of the feature variables in the case of the same training samples. Whereas the feature variables selected in the comparative experiments could only learn the spectral characteristics of water, it was difficult to accurately distinguish building shadows with similar spectral characteristics of water. Besides, the RF performed instable in different cities. For example, the false detection of the RF was more obvious in Guangzhou, but there were also obvious missing detections in the extraction of urban water in Kunming and Wuhan, some of which even occurred in the center of water. This indicated that the RF was not robust enough in water extraction for different areas, and was more suitable for urban water extraction at regional scale. Moreover, MNDWI and RF also had obvious errors around lakes within the central area of Wuhan. Most of these areas were vegetated farmland or small puddles that were mixed with other objects due to the spatial resolution of Landsat. MNDWI and RF could identify them as waters, indicating that these two methods were too sensitive to water signals.

Relatively speaking, in the four test cities, the OTOP method had performed well in both false detection and missing detection, both of which were within acceptable limits. This was because the MSCNN model trained offline not only relied on the spectral information of water, but also considered the spatial information, so it could determine the extents of urban water more accurately. In addition, since there were fewer water areas in Changchun, the extraction results of the RF and the OTOP method were similar visually. But closer observation showed that there were still a few false detections in the results of RF in urban centers. Compared with the RF, the OTOP method did improve the accuracy of water extraction in the building areas.

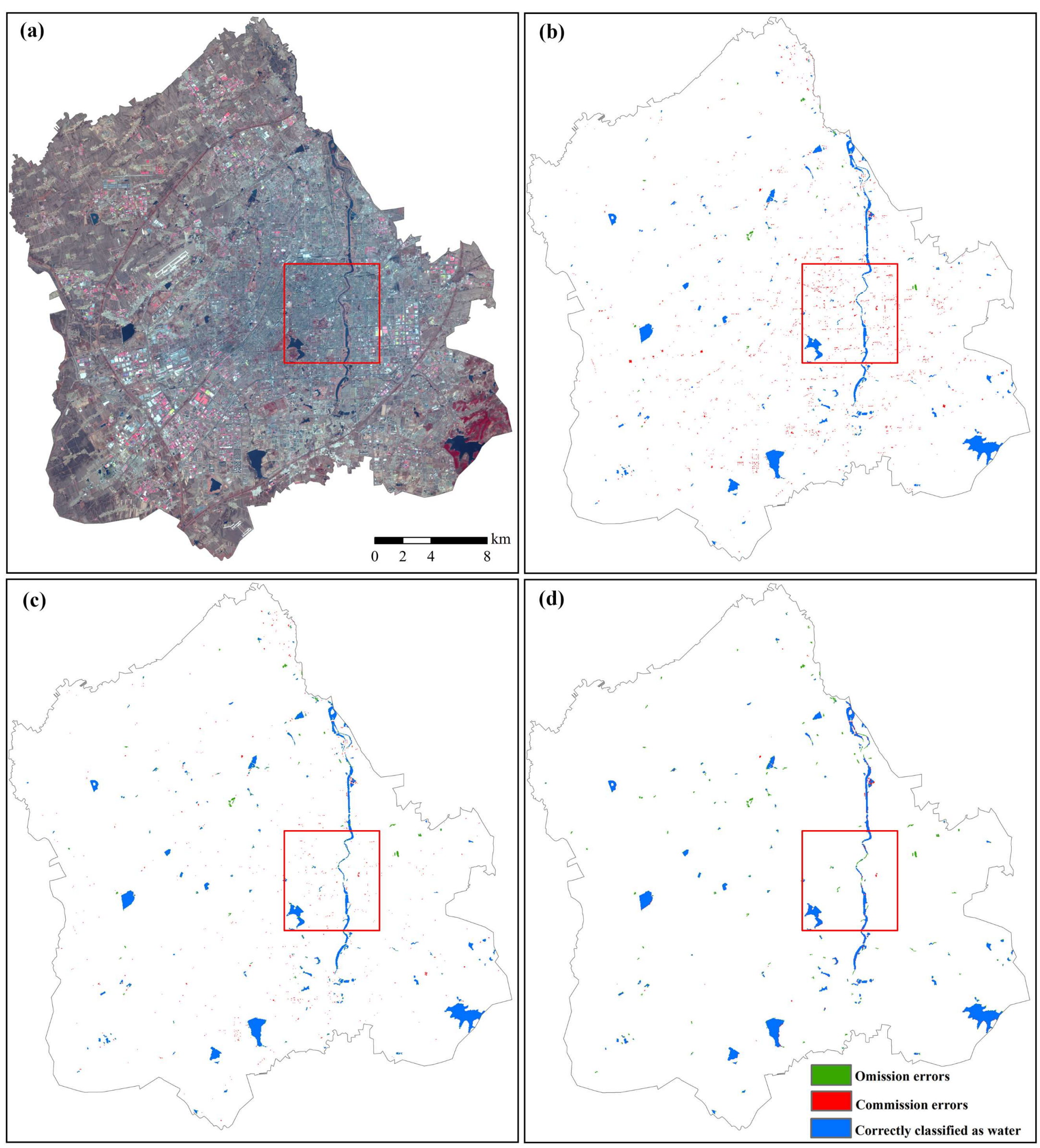

Fig. 3. The comparison of urban water extraction results in Changchun. (a) is the NIR-R-G false color display of the original Landsat image of the central urban area of Changchun. (b), (c) and (d) are the results of urban water mapping with MNDWI, the RF and the OTOP method, respectively. The blue in (b) - (d) shows the accurate detected urban waters, green shows the missing urban waters, and red shows the false detection. The colors in Fig. 4-7 represent the same meanings.

*2)* Urban water extraction accuracy

The confusion matrix could be obtained by comparing the predicted results and the reference masks pixel by pixel. Five indicators were calculated based on the confusion matrix to quantitatively evaluate the accuracy of urban water extraction, that is omission error (OE), commission error (CE), kappa coefficient (Kappa), F1-score and intersection over union (IoU, (2)). Note that kappa, F1-score and IoU are comprehensive indicators of the overall performance evaluation of a certain method, and a larger value means a higher accuracy, whereas the smaller the OE and CE are, the better.

$$IoU=\frac{\textit{Intersection areas of predicted and reference water}}{\textit{Union areas of predicted and reference water}} \quad (2)$$

The accuracies of MNDWI, RF and the OTOP method for extracting urban water in four test cities are summarized in Table II. The statistical results show that among the three methods, the OTOP method achieved higher accuracy in water extraction at all test cities. The mean kappa, mean F1-score and mean IoU of urban water extraction with the OTOP method in the four cities reached 0.924, 0.930 and 0.869, respectively, which were the highest in the comparison methods. Consistent with the results of the visual inspection, the MNDWI showed overestimation in the four test cities, that is, the false detection was serious. The omission errors of MNDWI in Guangzhou and Wuhan were even less than 1%, but the commission errors were around 20%, so the three comprehensive indicators were lower. In Kunming where the area with dense buildings was relatively small, the false detection of MNDWI was slightly better, the commission error was 13.06%, and the omission error was only 1.86%, the three comprehensive indicators were also relatively large. Compared with MNDWI, the accuracies of the RF had a certain degree of improvement. Except for Kunming, the overall accuracy of urban water extraction with RF in the other three cities had improved, and the commission error of each city had decreased, but the omission error had increased inevitably. The omission errors of Changchun and Kunming, which were 14.67% and 11.82%, increased the most. Whereas the commission errors of RF were larger in Guangzhou and Wuhan with the complex urban surface, which is 10.41% and 13.44%, respectively. On the whole, the quantitative evaluation results of RF also showed that although the overall accuracy of urban water extraction was improved, the extraction results in different cities were quite different. The result of water extraction of the RF in Kunming was not even as good as that of MNDWI. Therefore, the RF method is not stable enough for automated studies of large scale and long-term. Compared with these methods, the OTOP method performed better at all test cities. In addition to the obvious increase of the three comprehensive indicators, the omission errors and the commission errors of the four cities were also lower values, mostly below 10%. The only larger error was the omission error of Changchun, which was 14.38%. But compared with RF, it could be seen that the OTOP method reduced the commission error while making the omission error as far as possible, because the commission error and the omission error were mutually constrained. The results of other cities also showed that the OTOP method could achieve a better relative balance between the two indicators of commission error and omission error. That is to say, the OTOP method not only reduced the problem of commission errors in urban areas, but also guaranteed as much as possible that there were not many omission errors, so as to achieve satisfactory classification accuracy.

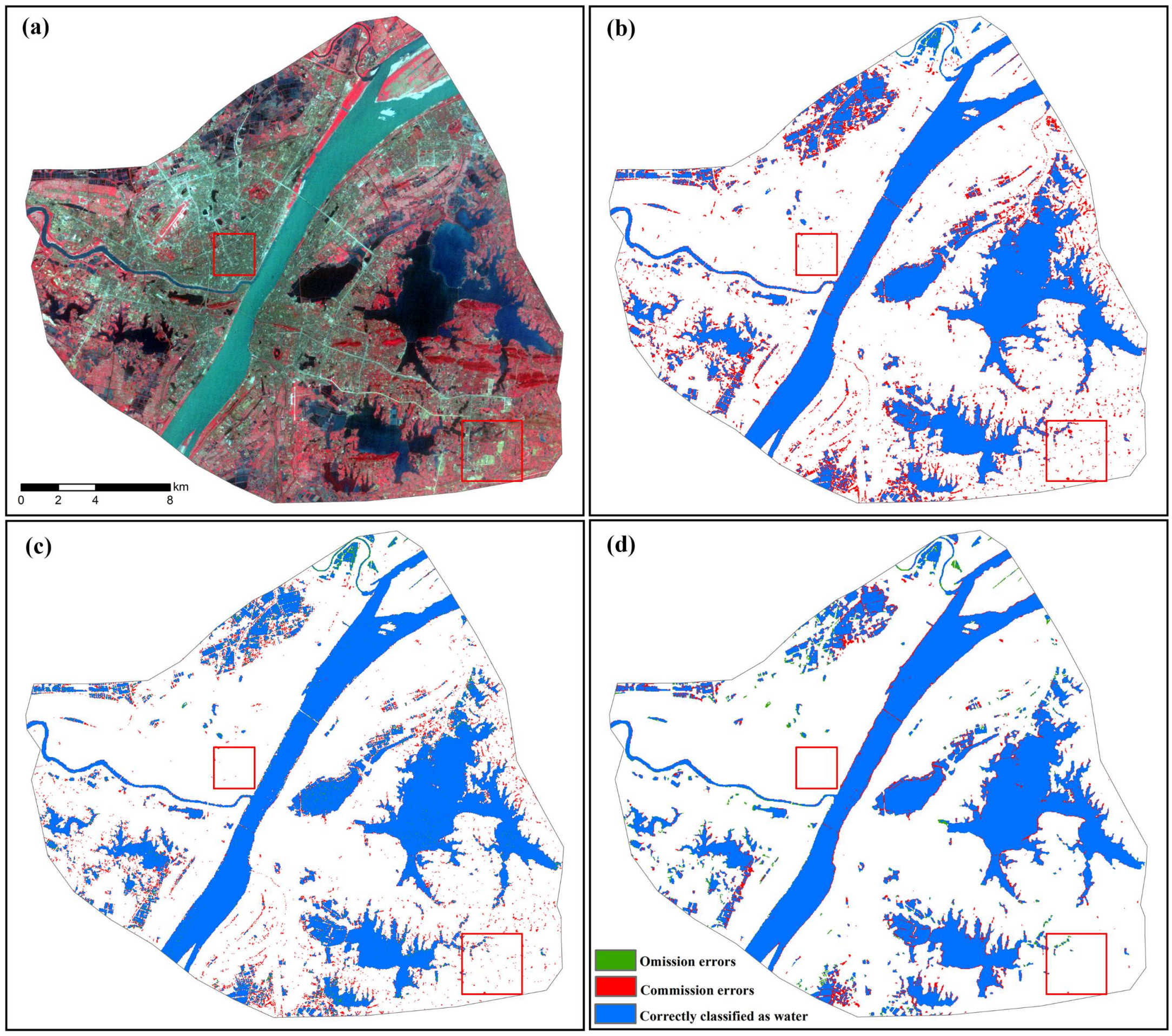

Fig. 4. The comparison of urban water extraction results in Wuhan. (a) is the NIR-R-G false color display of the original Landsat image of the central urban area of Wuhan. (b), (c) and (d) are the results of urban water mapping with MNDWI, the RF and the OTOP method, respectively.

Furthermore, from the perspective of overall performance,

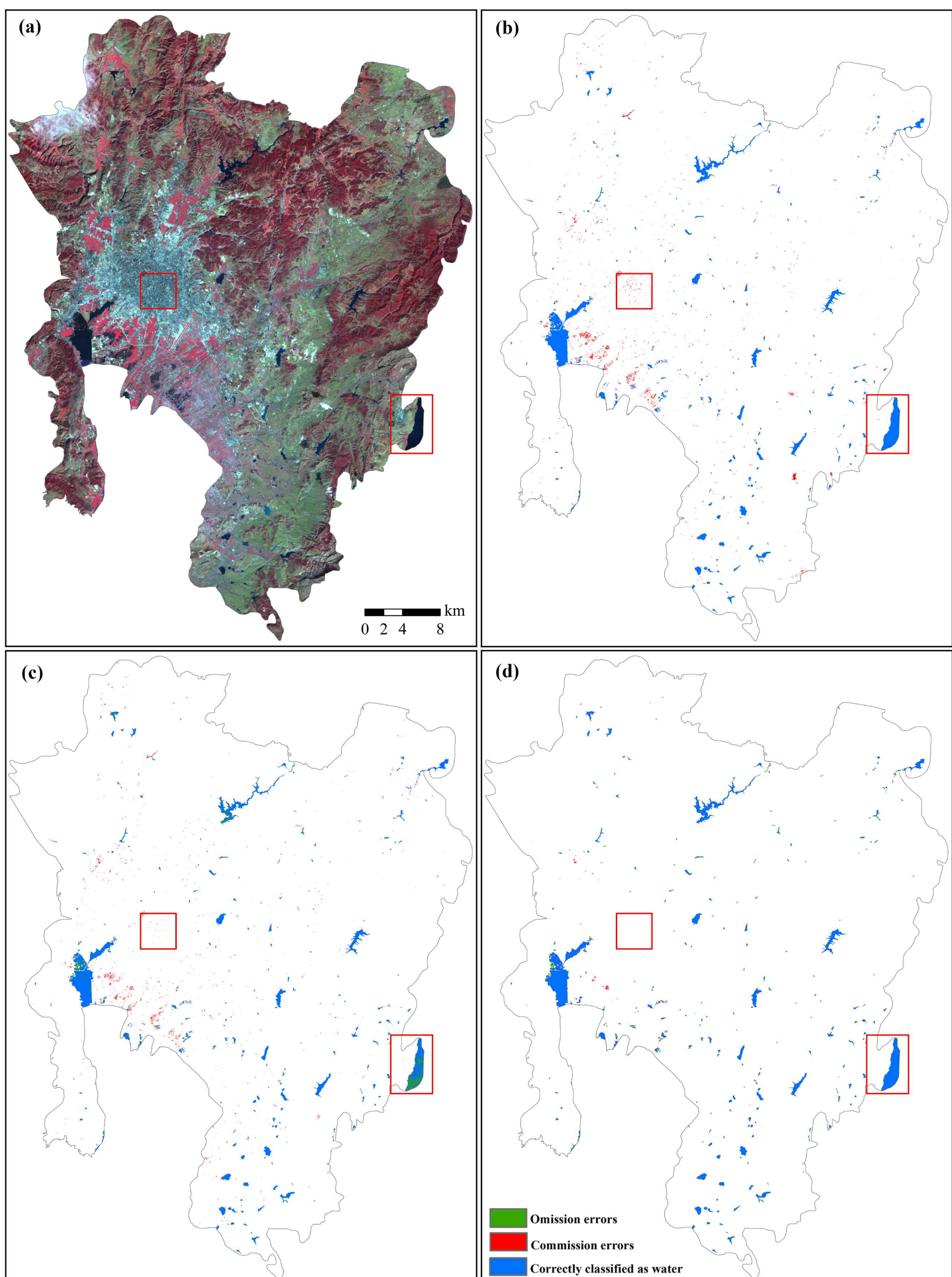

Fig. 5. The comparison of urban water extraction results in Kunming. (a) is the NIR-R-G false color display of the original Landsat image of the central urban area of Kunming. (b), (c) and (d) are the results of urban water mapping with MNDWI, the RF and the OTOP method, respectively.

the OTOP method had the least difference compared with the other two methods in images covering different test cities and from different sensors. The IoUs of the OTOP method were stable at higher values, and the kappa and F1-score were both above 0.9. This also showed that the OTOP method had strong robustness, and could accurately distinguish between water and non-water in urban areas.

## *B. Extended validation in major cities of China*

To further demonstrate the universality of the OTOP method, we extended the OTOP to the other 32 major cities in China, and evaluated the accuracies. According to the spatial distribution and development level of each city, the 32 cities were divided into eight groups without repetition, including four cities in each group. And there were nine groups in total with the four cities previously used for validation. The Landsat images and water masks of any eight groups were used as training data to extract urban water in the central urban areas of the remaining four cities. The above operation was repeated eight times, to complete the extended validation of the OTOP method in China. Then the same five indicators were also used to quantitatively assess the accuracy of urban water extraction in each city (Fig. 8).

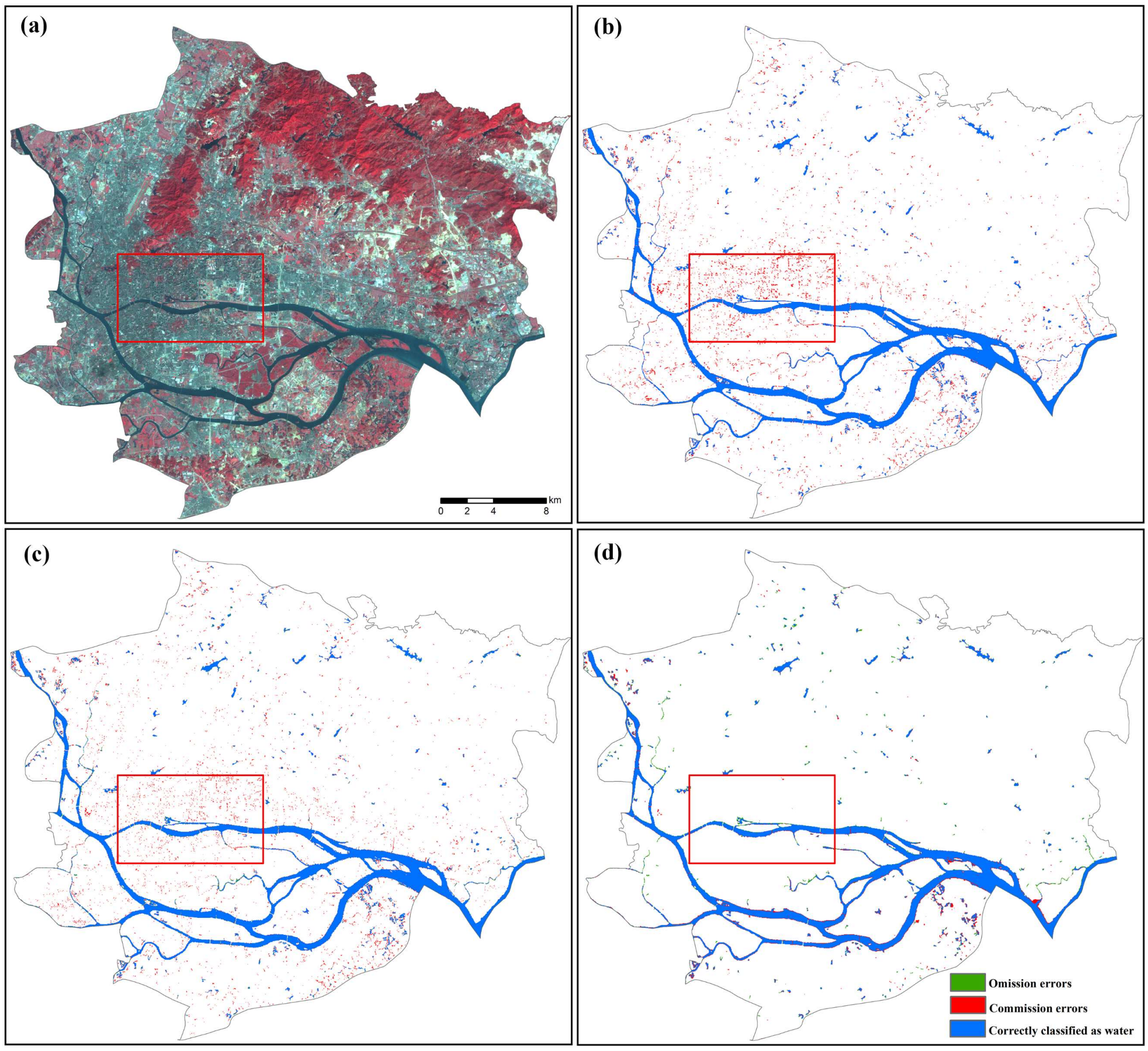

Fig. 6. The comparison of urban water extraction results in Guangzhou. (a) is the NIR-R-G false color display of the original Landsat image of the central urban area of Guangzhou. (b), (c) and (d) are the results of urban water mapping with MNDWI, the RF and the OTOP method, respectively.

TABLE II
THE ACCURACY ASSESSMENT RESULTS OF URBAN WATER EXTRACTION AT FOUR TEST CITIES BY MNDWI, RF AND THE OTOP METHOD.

| Test area | Method | OE | CE | Kappa | F1-score | IoU |
|---|---|---|---|---|---|---|
| Changchun | MNDWI | 6.87% | 20.38% | 0.856 | 0.859 | 0.752 |
| | RF | 14.67% | 8.37% | 0.882 | 0.884 | 0.792 |
| | OTOP | 14.38% | 5.30% | **0.901** | **0.900** | **0.817** |
| Wuhan | MNDWI | 0.26% | 20.57% | 0.843 | 0.884 | 0.793 |
| | RF | 8.45% | 10.41% | 0.876 | 0.906 | 0.828 |
| | OTOP | 3.76% | 9.74% | **0.912** | **0.934** | **0.877** |
| Kunming | MNDWI | 1.86% | 13.06% | 0.920 | 0.922 | 0.855 |
| | RF | 11.82% | 6.51% | 0.906 | 0.908 | 0.831 |
| | OTOP | 8.85% | 4.43% | **0.932** | **0.933** | **0.875** |
| Guangzhou | MNDWI | 0.58% | 19.33% | 0.885 | 0.893 | 0.807 |
| | RF | 5.01% | 13.44% | 0.899 | 0.906 | 0.828 |
| | OTOP | 6.45% | 10.10% | **0.912** | **0.917** | **0.847** |
| Average of the four test areas | MNDWI | 0.79% | 19.21% | 0.883 | 0.889 | 0.802 |
| | RF | 9.54% | 10.59% | 0.893 | 0.899 | 0.817 |
| | OTOP | 5.18% | 8.74% | **0.924** | **0.930** | **0.869** |

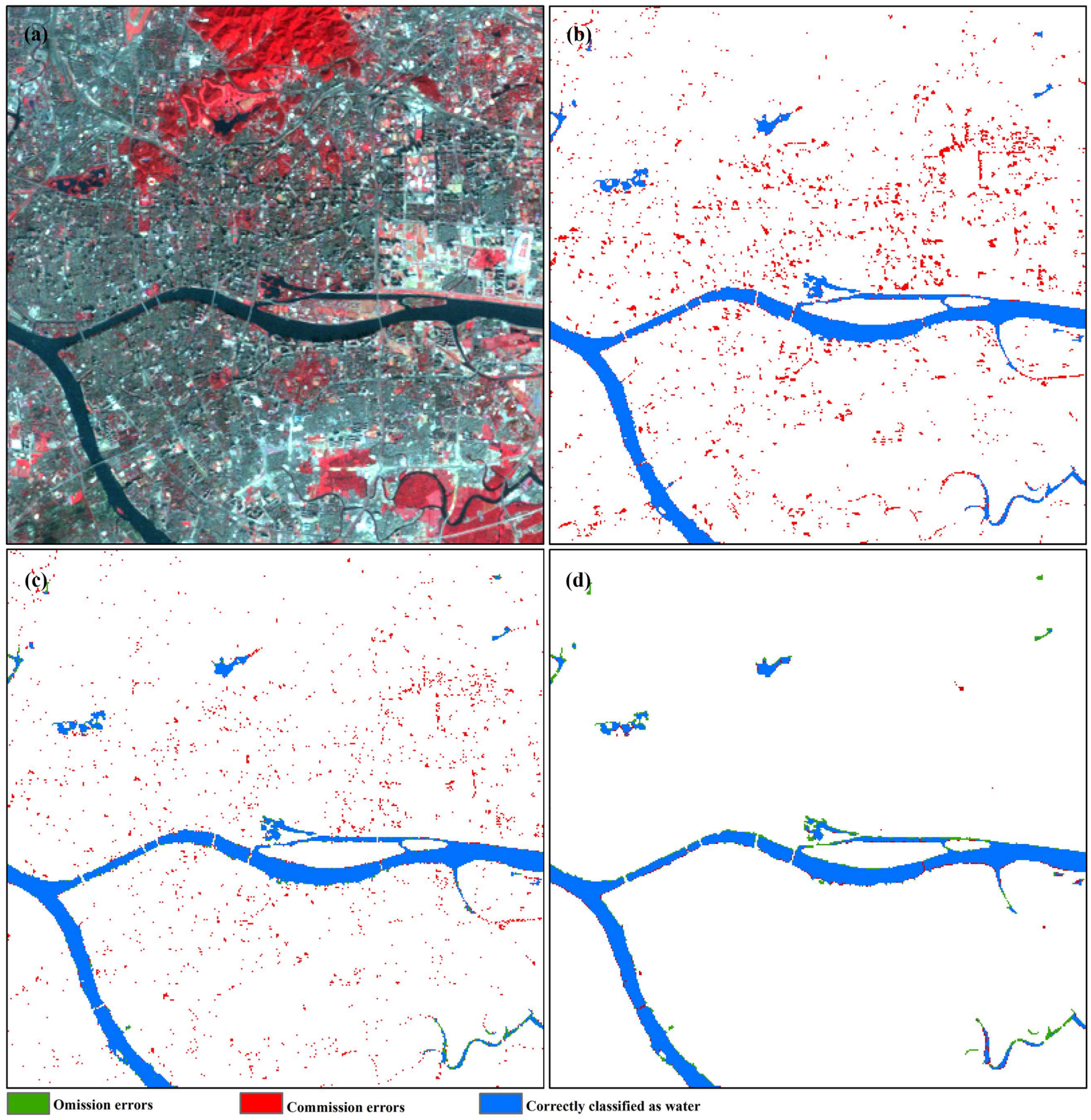

Fig. 7. The details of water extraction in urban dense building area of Guangzhou. (a) is the NIR-R-G false color display of the original Landsat image. (b), (c) and (d) are the results of urban water mapping with MNDWI, the RF and the OTOP method, respectively.

The results of extended validation showed that the OTOP method performed well throughout China and had a strong universality. The kappa, F1-score and IoU of most cities were above 0.9, 0.9 and 0.82, respectively. The OE and CE of urban water extraction in each city could also reach a satisfactory relative balance under the premise of mutual checks and balances. The OEs and CEs of most cities were below 13% and 10%, respectively. In the developed cities in eastern China, such as Beijing, Shanghai, Guangzhou and other cities with dense high-rise buildings, the OTOP method could accurately distinguish urban water and building shadows, and achieve higher water extraction accuracy. The OTOP method also had good performance in urban water extraction in some coastal cities such as Dalian, Qingdao, Fuzhou, Xiamen, Haikou, Taipei and so on. In some inland cities with relatively developed water systems such as Nanchang, Wuhan, Changsha, Chongqing, the kappa of the OTOP method were 0.949, 0.910, 0.941 and 0.980, respectively, and the IoUs were 0.910, 0.872, 0.892 and 0.962, respectively, which were all very high values in the accuracy evaluation. Their OEs and CEs of the OTOP method were also extremely small values. Whereas in some cities in western China, the comprehensive accuracies of water extraction were relatively lower, and the OEs were larger, such as Lhasa, Xining, and Hohhot, etc. There was less water and mainly slender rivers in these cities, such as the Lhasa River in Lhasa, the Huangshui River in Xining and the East River in Hohhot, so that there would be more mixed pixels in the water areas. And the water features of these mixed pixels were not obvious enough to be easily detected, but they were labeled as water according to the subjective consciousness of actual experience when manually drawing the water masks. Besides the influence of mixed pixels, too little water could also easily magnify the error indicators in accuracy assessment. Furthermore, it also can find that there were few false detections (small CEs) of urban water extraction in all cities. This also proved that the OTOP method could effectively distinguish water from building shadows and dark ground surfaces in different urban areas, and accurately determine the extents of urban water.

## V. Discussions

### A. Advantages of the method combining GEE and MSCNN

In this paper, we proposed the OTOP method for urban water extraction. The OTOP can not only improve the accuracy of urban water extraction in various environmental backgrounds, but also provided a flexible way to accelerate the efficiency of water extraction, so as to facilitate the related research of long-term and large-scale urban water monitoring.

GEE provides a powerful data storage and data processing platform which has free access to Landsat data across the entire time series, as well as the ability to quickly batch process large numbers of images regardless of the time and space. Therefore, compared with the other methods of urban water extraction which need to download data for subsequent processing locally, the OTOP method can save a lot of time for downloading data and space for storing data. And the data processing can be performed in parallel on GEE, so it can also make a qualitative leap in processing efficiency.

Moreover, the classification accuracy of the OTOP method has obvious improved compared with the traditional water extraction methods which were already encapsulated on GEE, and there is no need for additional data or separate detection procedures to remove noises from shadows and dark surfaces. Our experimental data was selected from Landsat images of 36 cities in different regions of China, which were deliberately chosen to cover different urban water types and urban surface features. If there are some clouds/cloud shadows in the image, the OTOP method also can accurately distinguish the water from them (Fig. 9). Therefore, the OTOP method can achieve higher precision requirements than the existing traditional water extraction methods on GEE, and it has strong robustness for different types of water extraction in different urban environments. MNDWI and the RF used as comparative experiments may also achieve higher accuracy by adjusting thresholds or changing combinations of feature variables. But this will have strong subjective consciousness and need rich experience as well as a lot of time, and it is not easy to be extended to large-scale automated application.

Compared with the latest deep learning function implemented on GEE with the TensorFlow framework, the OTOP method also has its unique advantages. The realization of the existing deep learning function on GEE provides a very convenient way for scientific research, but it needs to rely on three intermediate platforms of Google storage, Colaboratory and AI platform. When the usage of Google Storage exceeds the specific limits, or when using Google AI platform to train models and obtain predictions in the cloud, the cost is the factor that must be considered for some studies. As for the proposed OTOP method, the training of the network is done offline, and the powerful computing performance of GEE is sufficient for the part needed to process online, so the whole process is free. In addition, the OTOP method is not limited by the TensorFlow framework for the programming language and building style of the network model, so it can be more flexibly and freely to build the preferred deep learning model. The previously trained deep learning models can also be applied directly on GEE without reimplementing on TensorFlow.

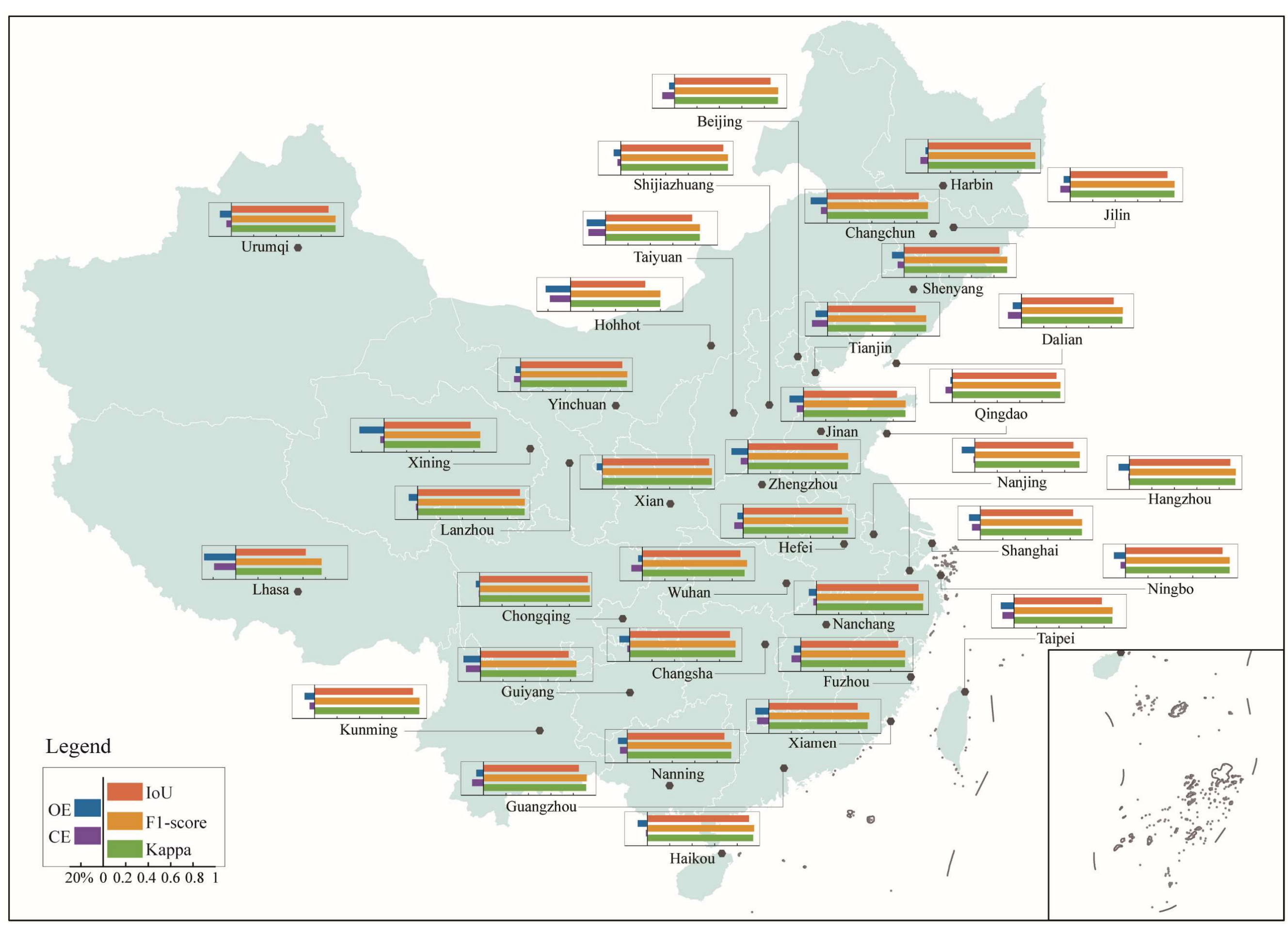

Fig. 8. The extended validation results of urban water extraction with the OTOP method in the major cities of China.

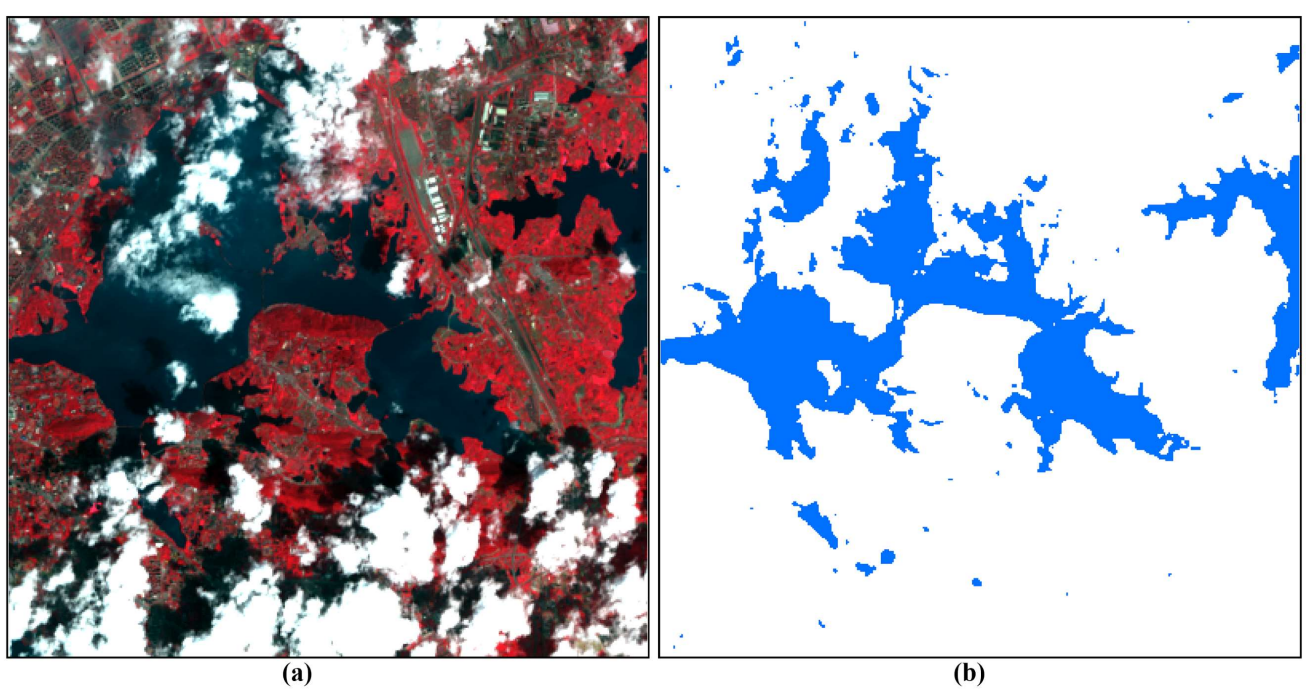

Fig. 9. The part of the water extraction result by the OTOP method for cloudy Landsat image with path/row of 123/039 on September 12, 2017. (a) is the NIR-R-G false color display of the original image, and the blue color in (b) indicates the extracted extents of water.

In a word, the combination of GEE and MSCNN does not need to make many subjective decisions affecting the accuracy of classification, but also maintains the advantage of high accuracy of urban water extraction using MSCNN. There is no need to take into account the download of massive data and the requirements for high-performance computing equipment which are necessary for the promoted applications. And we also provided a more flexible way to utilize deep learning models on GEE. The proposed OTOP method can be used to study the characteristics of urban water changes under the context of urbanization in China We can also apply the idea of the OTOP method to other studies of land use classification.

### B. *Further improvements*

Although the OTOP method achieved satisfactory results in this study, there are still several problems. Firstly, due to the limited spatial resolution of Landsat images, most of the edge pixels of water cover a relatively large area which may be composed of water and non-water objects [52]. The OTOP method tended to classify the mixed pixels at the edges of larger water areas as water during urban water extraction, which was one of the major reasons for the slight misdetection of water pixels in the predicted results (Fig. 4, Fig. 6). This problem may be solved by adding more accurate samples of edge pixels to the training data. In addition, because of the characteristic of convolution operations, the OTOP method is slightly inferior to the pixel-based water extraction method in terms of the fineness of water detection. Finally, the training data we selected for the offline training of MSCNN model was the artificially drawn water masks of Landsat images in China, so the model parameters may be more suitable for the mapping of urban water in Chinese cities. But there are more urban areas with other urban surface characteristics in the world. If we want to make the OTOP method applicable globally, it may need to include more training data in different urban areas.

## VI. CONCLUSION

Urban water mapping is very important for the urban management and planning. In this paper, we propose a new method combining GEE with CNN for urban water extraction, which is called the OTOP method. The method is to train a complete MSCNN model offline and then extract water in urban areas on GEE by using the trained model parameters. Such combination can not only give full play to the advantages of GEE specifically designed to manage big data, but also provide a more flexible way to use deep learning models on GEE to improve the accuracy of water extraction. In addition, the OTOP method only requires the six bands of Landsat images as input to separate water and non-water, and the multiscale feature fusion module in MSCNN can help to keep more information while considering the context during pixel by pixel classification. There is no need to manually select thresholds or join other artificially defined rules in different regions and conditions, nor to rely on rich prior knowledge. The results of comparative experiments with traditional MNDWI and RF method on GEE in Changchun, Wuhan, Kunming and Guangzhou also show that the OTOP method is better than other methods. It can effectively suppress the noises from building shadows and other dark surfaces in urban areas while ensuring accurate detection of urban water. In the extended validation of the other 32 major cities in China, the high accuracies of urban water extraction with the OTOP method suggest that it also has strong universality, and is suitable for water extraction at different times in different urban areas. Therefore, the OTOP method can meet the requirements for accuracy, automation level and wide application.

## ACKNOWLEDGMENT

The authors appreciate the editors and anonymous reviewers for their valuable suggestions.